\documentclass[runningheads]{llncs}
\usepackage[T1]{fontenc}
\usepackage{graphicx}
\usepackage{fancyhdr}
\usepackage{url} 
\usepackage{amsmath, amssymb} 
\usepackage{lipsum} 

\fancypagestyle{firstpageheader}{
    \fancyhf{} 
    \fancyhead[C]{\footnotesize PREPRINT: This is a preprint of the following paper, accepted by Springer for publication in the Communications in Computer and Information Science (CCIS) series.}.}

\begin{document}
\title{LLM as a Code Generator in Agile Model-Driven Development}
\author{Ahmed R. Sadik\inst{1}\orcidID{0000-0001-8291-2211} \and
Sebastian Brulin\inst{1}\orcidID{0000-0002-9710-6877} \and
Markus Olhofer\inst{1}\orcidID{0000-0002-3062-3829}}
\authorrunning{A. Sadik et al.}
%
\institute{Honda Research Institute Europe, Carl-Legien-Strasse 30, Offenbach am Main, Germany\\
\email{\{ahmed.sadik, sebastian.brulin, markus.olhofer\}@honda-ri.de}}
\maketitle 

\thispagestyle{firstpageheader}

\begin{abstract}
Leveraging Large Language Models (LLM) like GPT-4 in the auto-generation of code represents a significant advancement, yet it is not without its challenges. The ambiguity inherent in natural language descriptions of software poses substantial obstacles to generating deployable, structured artifacts. This research champions Model-Driven Development (MDD) as a viable strategy to overcome these challenges, proposing an Agile Model-Driven Development (AMDD) approach that employs GPT-4 as a code generator. This approach enhances the flexibility and scalability of the code auto-generation process and offers agility that allows seamless adaptation to changes in models or deployment environments. We illustrate this by modeling a multi-agent Unmanned Vehicle Fleet (UVF) system using the Unified Modeling Language (UML), significantly reducing model ambiguity by integrating the Object Constraint Language (OCL) for code structure meta-modeling, and the FIPA ontology language for communication semantics meta-modeling. Applying GPT-4's auto-generation capabilities yields Java and Python code that is compatible with the JADE and PADE frameworks, respectively. Our thorough evaluation of the auto-generated code verifies its alignment with expected behaviors and identifies enhancements in agent interactions. Structurally, we assessed the complexity of code derived from a model constrained solely by OCL meta-models, against that influenced by both OCL and FIPA-ontology meta-models. The results indicate that the ontology-constrained meta-model produces inherently more complex code, yet its cyclomatic complexity remains within manageable levels, suggesting that additional meta-model constraints can be incorporated without exceeding the high-risk threshold for complexity.
\end{abstract}
\section{Introduction}
In the era of artificial intelligence, with Large Language Models (LLMs) trained on diverse codebases, new opportunities for innovation in Model-Driven Development (MDD) emerge. MDD seeks to enhance the efficiency and durability of software engineering practices. Accordingly, this work extends our research in \cite{sadik2024coding} to augment Agile Model-Driven Development (AMDD). Our approach leverages existing LLMs, such as GPT-4, to auto-generate deployment-ready software artifacts \cite{sadik2023analysis}. Our objective is to conserve the substantial time and effort traditionally required in MDD for developing and updating a unique code generator for each deployment \cite{hailpern2006model}, ensuring that the auto-generated code is well-structured and meets its intended functionality and specified requirements \cite{feltus2017modeldriven}.

Code generation from formal graphical languages like the Unified Modeling Language (UML), Systems Modeling Language (SysML), or Business Process Model and Notation (BPMN) diagrams has become a cornerstone of contemporary software engineering \cite{omg2006object}. With appropriate tools, these languages can be seamlessly translated into executable code, enhancing the efficiency of the software development process \cite{sharaf2019modeling}. By integrating static and dynamic aspects from these languages and refining the model with declarative languages such as the Object Constraint Language (OCL) \cite{cabot2012object} or FIPA meta-ontology \cite{fipa2000ontology}, the generation of complex, well-structured code is facilitated \cite{kapferer2020domain,perezmartinez2004uml}. This automation ensures alignment between design and implementation, minimizes coding errors, and elevates software quality, thereby accelerating market entry \cite{sarkisian2022system}. A deep understanding of a system's dynamic behavior and overarching views is crucial for grasping its functionality \cite{siricharoen2009ontology}.

Assessing the quality of auto-generated code is imperative in validating our proposed AMDD approach. Traditional measures of testability, maintainability, and reliability, being qualitative, were susceptible to subjective bias \cite{liu2023improving}. Our study, however, aims to apply objective, measurable standards. At the heart of our analysis is the structural integrity of the code generated from models, for which we employed cyclomatic complexity as a key measure of structural soundness. Furthermore, our evaluation encompasses a comparative analysis of the code's functionality across different programming languages, against the expected behaviors delineated in the model's design \cite{ahmad2023towards}.

Section 2 of this article outlines the problem statement, spotlighting the inflexibilities of current MDD practices. Section 3 details the proposed AMDD approach. Section 4 introduces the UML model of a case study on an Unmanned Vehicle Fleet (UVF), and Section 5 integrates the case study's structure and communication meta-models using OCL and FIPA ontology language, respectively. Section 6 delves into the cyclomatic complexity of the generated code to evaluate its structure and compares the behavior of two deployments in Java and Python to assess its functionality. Section 7 wraps up with our principal findings, their implications, and directions for future research.

\section{Problem Statement}

Natural language's inherent ambiguity presents challenges not only for machine interpretation but also for human understanding. In leveraging GPT-4 in deployable code auto-generation, the unclear and open-ended nature of the input prompts frequently results in flawed and incomplete code. This problem is especially noticeable in scenarios where the software to be generated composed of different interconnected artifacts. Thus, rendering the software is impossible to be precisely encapsulated in natural language. MDD provides a solution to this problem by offering high-level models as the primary source for generating final code. However, designing and maintaining these models poses challenges that can detract from MDD's efficacy. First, while traditional modeling methods like UML excel at structuring data, they often fall short in providing the semantic depth and rule-based knowledge \cite{belghiat2012from}. Consequently, the resulting models, though structurally sound and behaviorally accurate, lack the semantic richness required for effectively representing complex real-world situations. Second, converting these models into executable code is far from direct \cite{petrovic2023automated}. It necessitates manual crafting and continuous updating of the code generator to align with model modifications and evolving technology stacks, a challenge amplified when switching deployment across different programming languages \cite{camara2023assessment}.

\begin{figure}[h!]
\centering
\includegraphics[width=6cm]{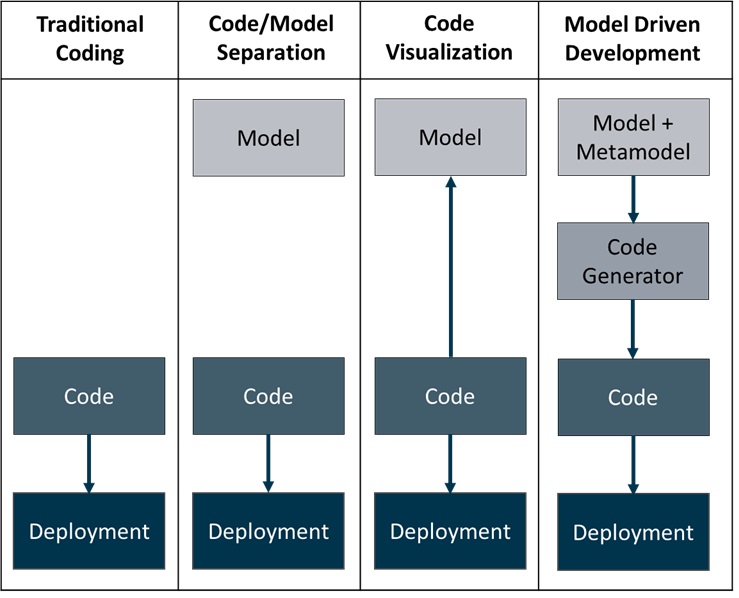}
\caption{Traditional coding vs MDD \cite{sadik2024coding}.} 
\label{fig1}
\end{figure}

Addressing the challenges in code generation within the existing  MDD framework necessitates a clear understanding of the distinctions between traditional coding practices and MDD code generation, as depicted in Fig.~\ref{fig1}. Traditional coding involves directly translating software functionalities into code, which suits smaller, straightforward features. Debugging, testing, and maintenance occur at the code level. Conversely, the model-and-code separation approach uses models to abstractly grasp the system, with models often discarded post-coding due to the prohibitive cost of updates. In contrast, MDD positions models as the cornerstone of the development process, substituting source code with source models from which the target code is auto-generated. This elevates the abstraction level and simplifies complexity. Tools such as Eclipse Papyrus, MagicDraw, Enterprise Architect, and IBM Rational Rhapsody have supported this process \cite{david2023blended}. However, any significant model updates or changes in the deployment language necessitate extensive modifications to the code generators, hindering development agility. Crafting a code generator for each programming language is both time-consuming and complex, limiting the MDD approach's adaptability and agility across different languages.

Our study explores leveraging LLM like GPT-4 as universal code generators to address these limitations. Despite MDD offering a structured approach that surpasses natural language's inadequacies for generating deployable code, the current MDD methodology hasn't fully embraced the capabilities of modern LLMs in auto-generating code. This gap renders the traditional MDD approach less effective within agile software development workflows, highlighting the need for alignment with contemporary LLM capabilities in code generation.

\section{Proposed Approach}

\begin{figure}[h!]
\centering
\includegraphics[width=8cm]{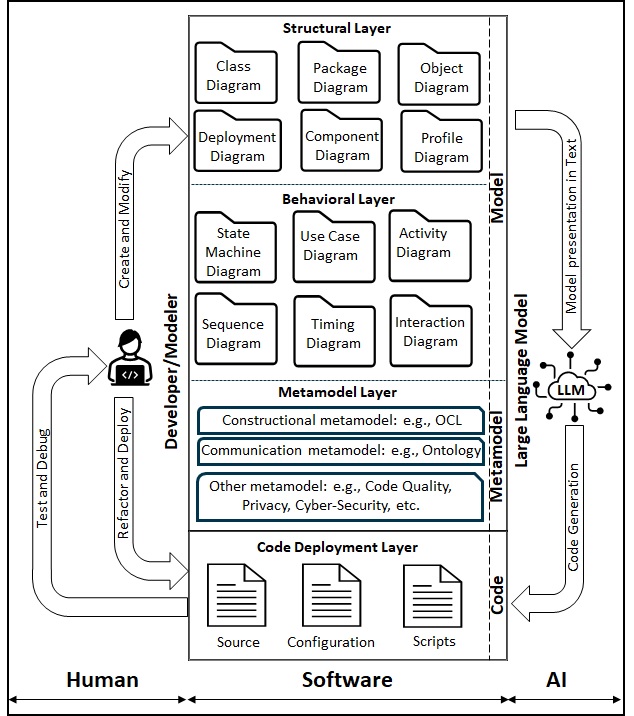}
\caption{Proposed AMDD approach \cite{sadik2024coding}.}
\label{fig2}
\end{figure}

Addressing the challenges described in the problem statement, our AMDD approach uses GPT-4 as a code generator , that  seamlessly interprets of the model into interconnected software artifacts that are ready for deployment. We leveraged PlantUML to translate UML diagrams into a formal text representation, to bridge the gap between GPT-4 text-based prompt and the visual UML's diagrams, therefore facilitating direct input into GPT-4. As illustrated in Fig.~\ref{fig2}., the modeler starts by crafting the model's layers: structural, behavioral, and constraints.

The structural layer outlines the static aspects of the model, showcasing software components and their complex relationships. Class diagrams elucidate object relationships and hierarchies, package diagrams group objects to highlight dependencies, and component diagrams offer a high-level view of system functionality and inter-component links. Deployment diagrams illustrate the hardware setup and component distribution for the system's physical architecture, whereas object diagrams provide runtime snapshots of objects, and profile diagrams adapt UML models for specific platforms.

The behavioral layer depicts the system's operations and interactions through various diagrams. Sequence diagrams map out event sequences, offering a timeline of interactions, while activity diagrams provide a detailed process flow. Interaction diagrams and timing diagrams further explore component interplay and the significance of timing, respectively. Use case diagrams demonstrate interactions between external entities and the system, and state diagrams trace the lifecycle and state transitions of entities.

Despite the comprehensive view offered by structural and behavioral diagrams, they fall short in articulating the governing rules and semantics. Our research introduces a constraints layer to refine the model's architecture by incorporating explicit meta-values and rules beyond UML's scope. We employ the OCL to define detailed code construction rules for both layers, specifying class invariants, method pre- and post-conditions, and parameter value restrictions. Additionally, communication constraints are articulated using ontology languages, enhancing knowledge sharing among software artifacts.

In the deployment layer, we utilize GPT-4 for its advanced reasoning capabilities, crucial for integrating model semantics into the auto-generated code. After ChatGPT generates the code, the modeler is responsible for deploying and validating its functionality. While ChatGPT's code generation is a powerful tool, it's evolving and not without errors \cite{dong2023selfcollaboration}. Anticipating and addressing potential deployment bugs, possibly with ChatGPT's assistance, is essential for ensuring the code fulfills its intended function effectively.

\section{Case study model}

The use-case selected for discussion centers around an UVF, consisting of a variety of UVs tasked with distinct missions, all orchestrated by a Mission Control Center (MCC) with human oversight. This case study has been designed as inherently distributed, making it suitable for representation and analysis through a Multi-Agent System (MAS) \cite{brulin2023bilevel}. The case study complexity is notably high since each participating entity is modeled as an agent, which necessitates ongoing communication and information sharing among agents to fulfill a unified objective, namely the successful completion of the fleet's mission. To ensure clarity and prevent the details of the MAS from becoming overwhelming, subsequent sections will focus solely on the fundamental model, providing insights into the operational principles of the MAS without delving into its complexities.

\subsection{Structure model}
\begin{figure}[h!]
\centering
\includegraphics[width=10cm]{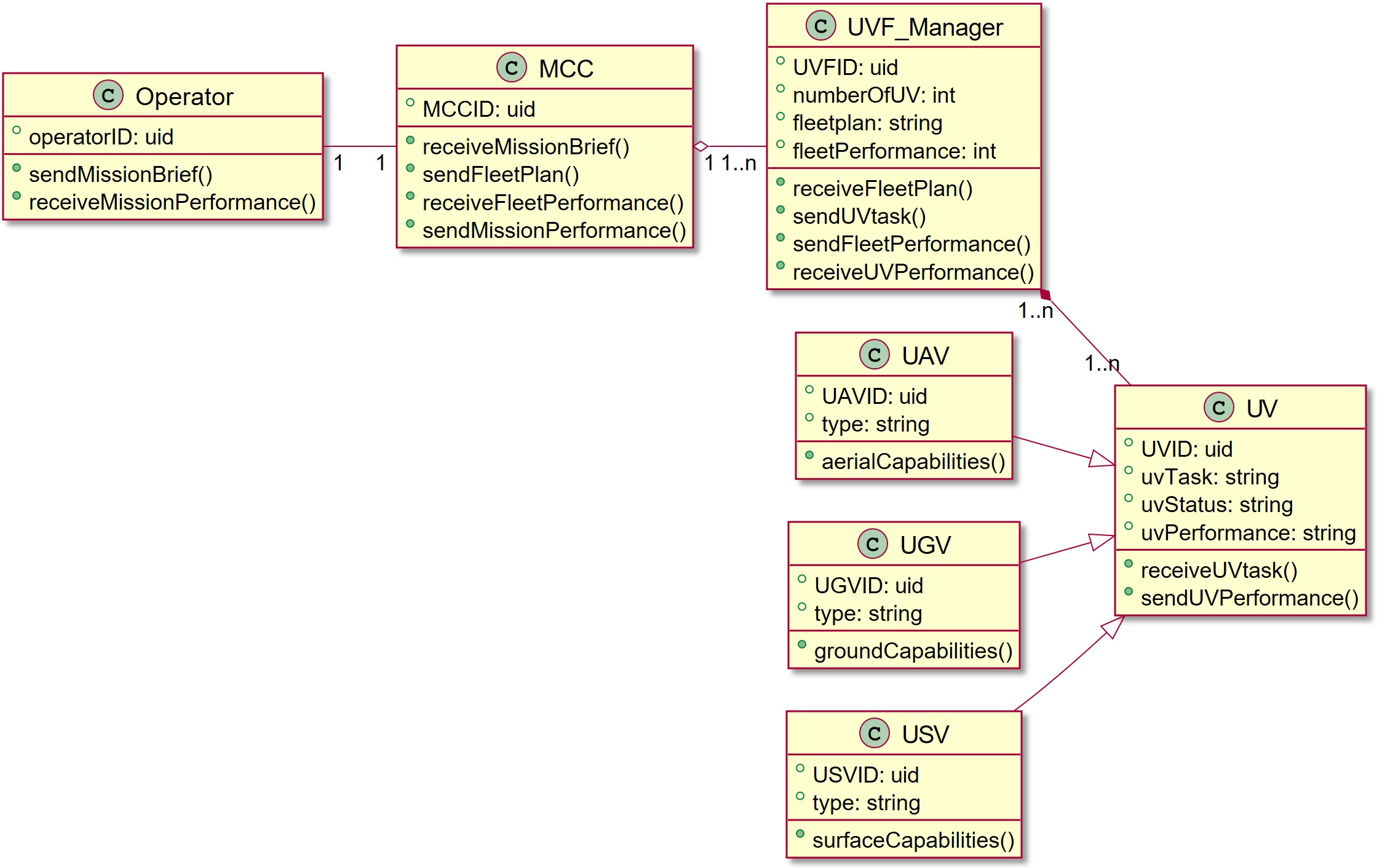}
\caption{Case-Study class diagram \cite{sadik2024coding}.} 
\label{fig3}
\end{figure}

Within the architecture of the model, the class diagram holds a crucial role by mapping out each participant in the case study as a distinct agent class, a relationship depicted in Fig.~\ref{fig3}. This diagram introduces the operator agent, embodying the human operator's functions such as dispatching mission briefings and collecting performance metrics. It also details the MCC agent, defined by characteristics like its unique MCC-ID, which oversees mission coordination and fleet supervision. Additionally, the UVF-Manager agent is tasked with the oversight of fleet operations, encompassing the allocation of tasks and monitoring of execution outcomes. The Unmanned Vehicle (UV) agent is conceptualized as a foundational class for UVs, branching into specific subclasses for varied vehicle categories, including UAVs, UGVs, and USVs . Through this class diagram, the intricate interactions among agents are clarified, highlighting their attributes, functionalities, and the web of relationships among them, such as composition, aggregation, and inheritance, while also defining the connections' cardinality.

\subsection{Behaviour model}

\begin{figure}[h!]
\centering
\includegraphics[width=8cm]{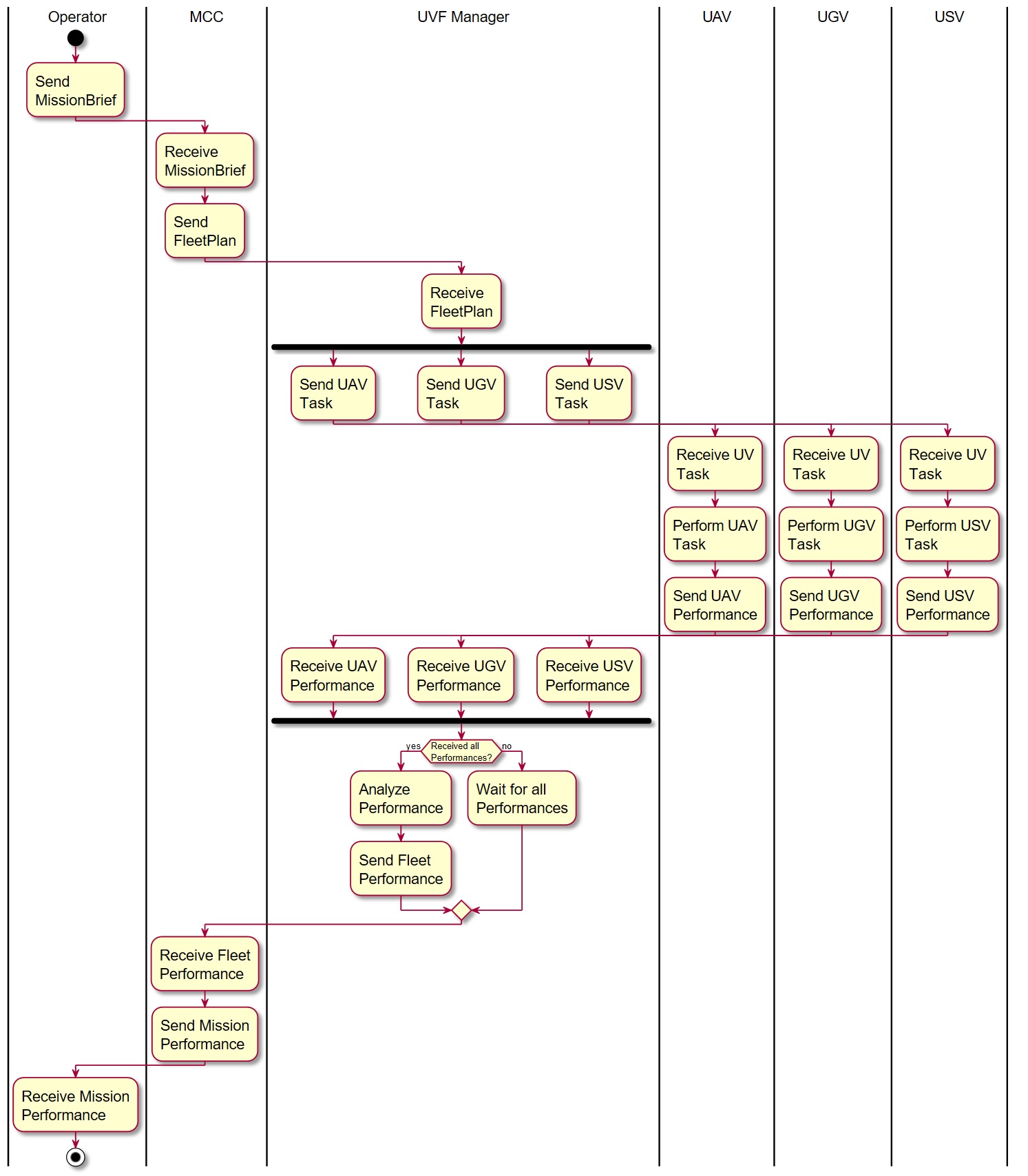}
\caption{Case-Study activity diagram \cite{sadik2024coding}.} 
\label{fig4}
\end{figure}

For conciseness and focus, the article will explore only two significant behavioral views critical for comprehending the case study model: the activity and state diagrams. These diagrams are essential for understanding the dynamic aspects of the case study. The activity diagram, detailed in Fig.~\ref{fig4}, enriches the class diagram by thoroughly describing processes such as synchronization, parallel execution, and conditional flows, vital for the successful completion of mission objectives. Conversely, the state diagram offers a detailed exploration of the agent classes' lifecycle within the model, revealing their coordination and reactions to achieve the overarching mission goals.

The activity diagram in Fig.~\ref{fig4} dissects the flow of information and tasks among the agents, illustrating the orchestration of processes and the sequence of task allocation, execution, and evaluation. This diagram highlights the temporal and logical dynamics of the MAS. The interaction commences when the operator agent forwards the mission brief to the MCC agent, which then decodes the brief into a strategic plan and relays it to the UVF-manager. This manager is responsible for task distribution to the available UVs. Upon completion of these tasks, the UVF-manager gathers the performance data from each UV, crucial for assessing the UVF's overall performance. This comprehensive performance feedback is sent back to the MCC, where it is translated into mission performance indicators and communicated back to the operator agent.

\begin{figure}[h!]
\centering
\includegraphics[width=6cm]{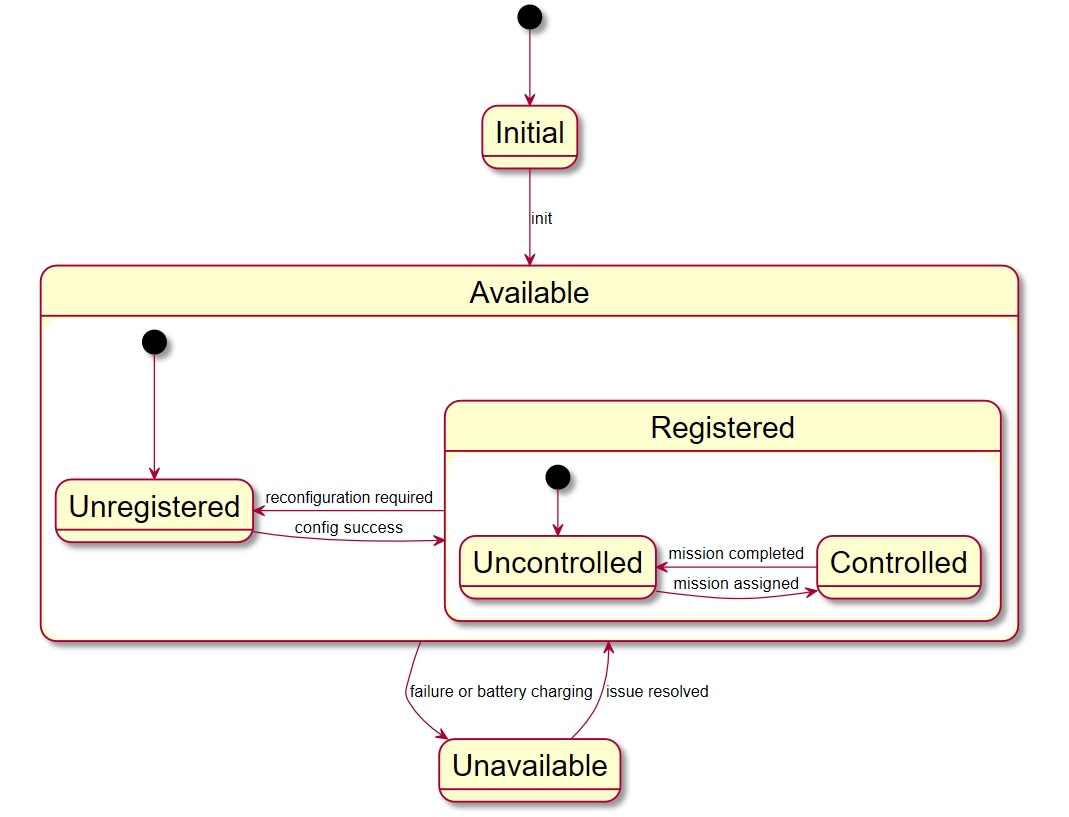}
\caption{Case-Study state diagram \cite{sadik2024coding}.}
\label{fig5}
\end{figure}

The state diagram further uncovers the internal workings of each agent, illustrating the transitions and actions that various events can initiate. For the operator, MCC, and UVF-manager, their statuses are straightforwardly represented by either 'busy' or 'free' states. The UVs, on the other hand, demand a nuanced state machine for a precise evaluation of task performance. As shown in Fig.~\ref{fig5}, the UV states are categorized as 'Available', indicating the UV is ready for tasks regardless of registration status; 'Unavailable', when a UV is out of service; 'Unregistered', ready for use but not yet enlisted; 'Registered', officially listed and potentially tasked, which further divides into 'Uncontrolled', listed but idle, and 'Controlled', actively assigned to a mission. This approach streamlines the understanding of each UV's current operational condition and its capacity to undertake mission-specific tasks.

\section{Case study metamodel}

In the proposed AMDD framework, the meta-model layer serves a pivotal role by encapsulating constraints that cover all facets of technical requirements not directly representable in the model layer. These constraints act as the guiding principles or rules that ensure the model adheres to specified technical standards and requirements. Upcoming sections will delve into the various types of meta-model constraints employed in the case study's modeling process, providing a comprehensive overview of how these constraints influence and shape the model's development and ensuring it aligns with the desired technical specifications and objectives

\subsection{Construction metamodel}

OCL complements the UML by defining rules for classes within a model. It plays a crucial role in adding construction constraints to UML diagrams, thereby enhancing model refinement and clarity. By addressing model ambiguities, OCL ensures precise modeling, which is particularly beneficial for generating deployed code directly from UML. This precision facilitates a smoother and more accurate transition from model to code.

\begin{figure}[h!]
\centering
\includegraphics[width=7cm]{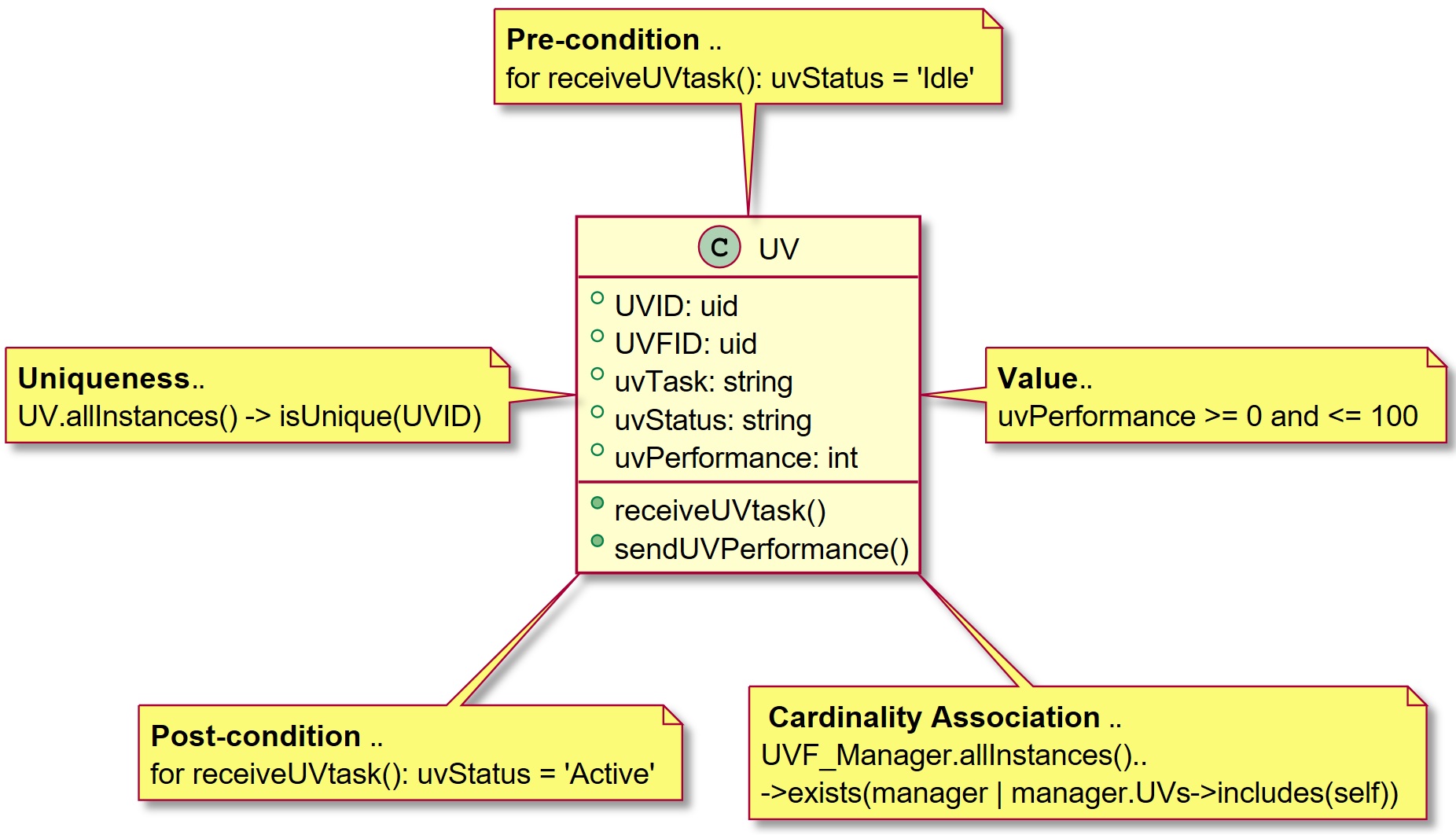}
\caption{UV agent construction metamodel example \cite{sadik2024coding}.} 
\label{fig6}
\end{figure}

Our research highlights the application of OCL through the implementation of five specific types of constraints on all agent classes, as illustrated with the UV agent class example in Fig.~\ref{fig6}. These constraints are:
\begin{itemize}
    \item \textbf{Uniqueness:} Ensures every agent has a distinct identifier.
    \item \textbf{Cardinality:} Manages relationships between agents, such as a UVF-manager to multiple UVs.
    \item \textbf{Value:} Limits class attributes within certain ranges, like UV performance scores between 0 and 100.
    \item \textbf{Pre-condition:} Verifies agents are in the correct state before transitions, for instance, a UV can only accept tasks if idle.
    \item \textbf{Post-condition:} Dictates state changes after transitions, such as updating a UV's status to 'Active' after task assignment.
\end{itemize}
These constraints significantly improve model precision and reliability by strictly adhering to defined rules and conditions.

\subsection{Communication metamodel}

While the OCL aims to define constraints for classes within UML models, it has limitations in facilitating inter-class communication—essential for the functionality of MAS ~\cite{sadik2017combining,sadik2016holonic}. To bridge this gap, technologies such as Java Agent DEvelopment (JADE) and Python Agent DEvelopment (PADE) turn to the FIPA-ontology communication language ~\cite{sadik2018cprosaholarchy}. This language significantly enhances MAS by providing a rich set of interaction protocols tailored for complex agent communications, overcoming the shortcomings of OCL in this domain. Our case study specifically points out these limitations of OCL and demonstrates how FIPA-ontology language serves as a crucial tool in developing sophisticated MAS communications.

\begin{figure}[h!]
\centering
\includegraphics[width=9cm]{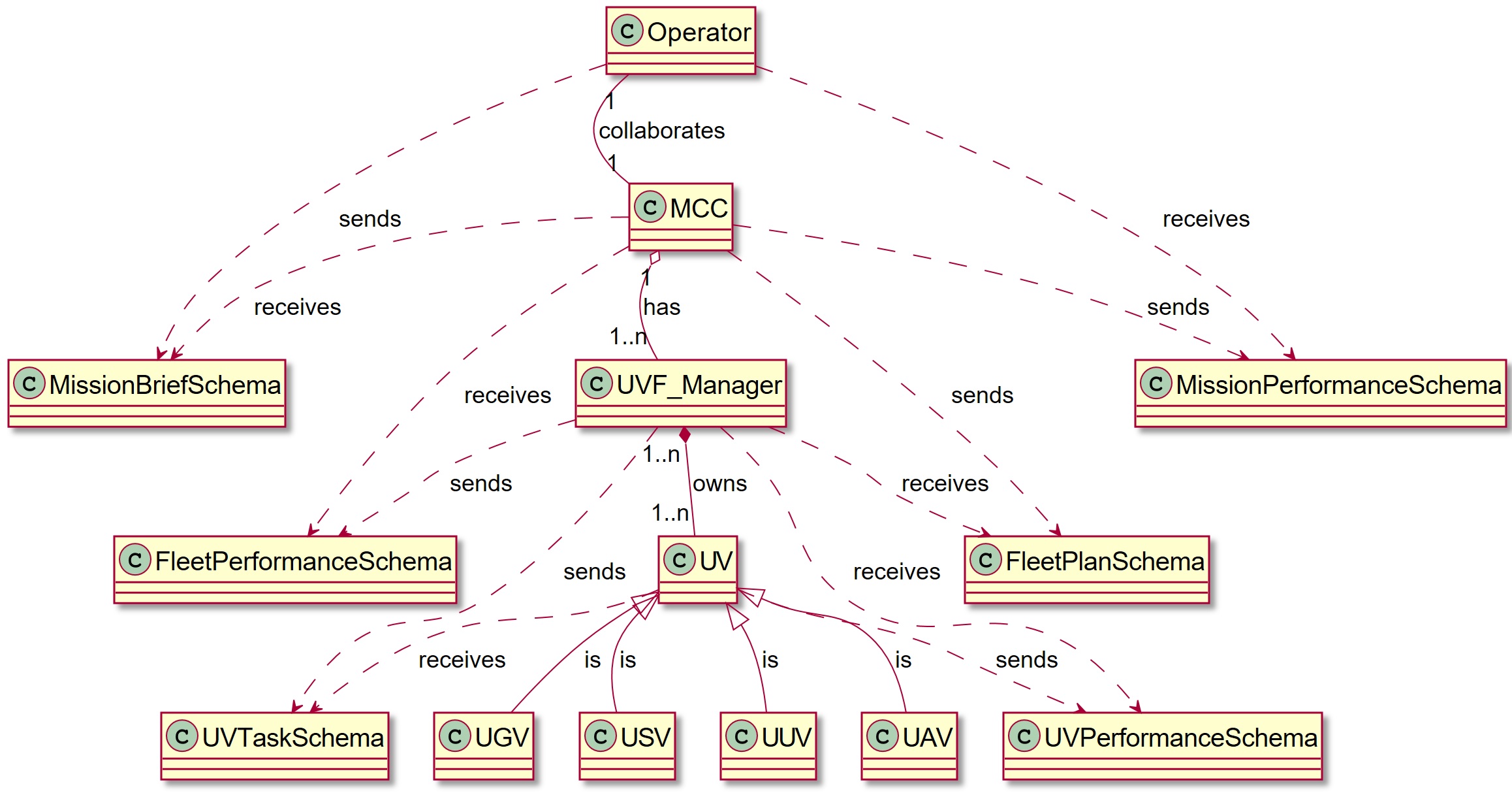}
\caption{UV agent construction metamodel example \cite{sadik2024coding}.} 
\label{fig7}
\end{figure}

In our MAS communication framework, as depicted in Fig.~\ref{fig7}, the FIPA-ontology establishes a series of structured schemas vital for facilitating complex communications within the system. These schemas are divided into key sets:

\textbf{Concepts:} These schemas form the backbone of mission and agent management communication, including:
\begin{itemize}
    \item \textit{Mission-Brief Schema:} Details such as mission-ID and status.
    \item \textit{Fleet-Plan Schema:} Specifies fleet configurations.
    \item \textit{UV-Task Schema:} Assigns tasks to individual unmanned vehicles (UVs).
    \item \textit{UV-Performance Schema:} Records performance metrics of UVs.
    \item \textit{Fleet-Performance} and \textit{Mission-Performance} schemas, each focusing on specific performance metrics.
\end{itemize}

\textbf{Predicates:} These schemas define the relationships between different agent classes, covering:
\begin{itemize}
    \item \textit{Inheritance:} E.g., a UAV as a subtype of UV.
    \item \textit{Composition:} Such as an MCC incorporating a UVF-manager.
    \item \textit{Aggregation:} As seen in a UVF-manager overseeing multiple UVs.
    \item \textit{Collaboration:} Demonstrated by the interaction between an operator and the MCC.
\end{itemize}

\textbf{Actions:} Operations that agents can execute, especially regarding message schemas, include:
\begin{itemize}
    \item Sending and receiving data, like an operator agent sending a mission brief to the MCC or the MCC receiving it from the operator.
\end{itemize}

This structured approach, as provided by the FIPA-ontology, not only clarifies the communication protocols within the MAS but also ensures that interactions are both efficient and effective, catering to the complex needs of agent-based communication.

\section{Code evaluation}

In the final stage of our AMDD approach, we employ GPT-4 for the conversion of the model and the metamodel into executable code. Our findings indicate an average occurrence of four bugs per agent class generated by GPT-4, predominantly due to the omission of necessary library imports. Nevertheless, with the correction of these bugs, the generated code becomes suitable for deployment. It is important to note that our primary interest in this study lies not in the accuracy of the auto-generated code but in its comprehensiveness. Our evaluation focuses on the examination and analysis of the structure and behavior of the auto-generated code, rather than on identifying and quantifying the bugs it may contain.

To this end, we conducted two distinct experiments. The first experiment sought to dissect the structure and complexity of the auto-generated code, providing insights into its architectural design and the intricacies of its internal mechanisms. The second experiment was designed to investigate the behavior of the auto-generated code, aiming to understand how well it performs its intended functions within a given context. These experiments collectively offer a holistic view of the auto-generated code's efficacy, highlighting areas of strength and opportunities for further refinement.

\subsection{Structure evaluation}

\begin{figure}[h!]
\centering
\includegraphics[width=2cm]{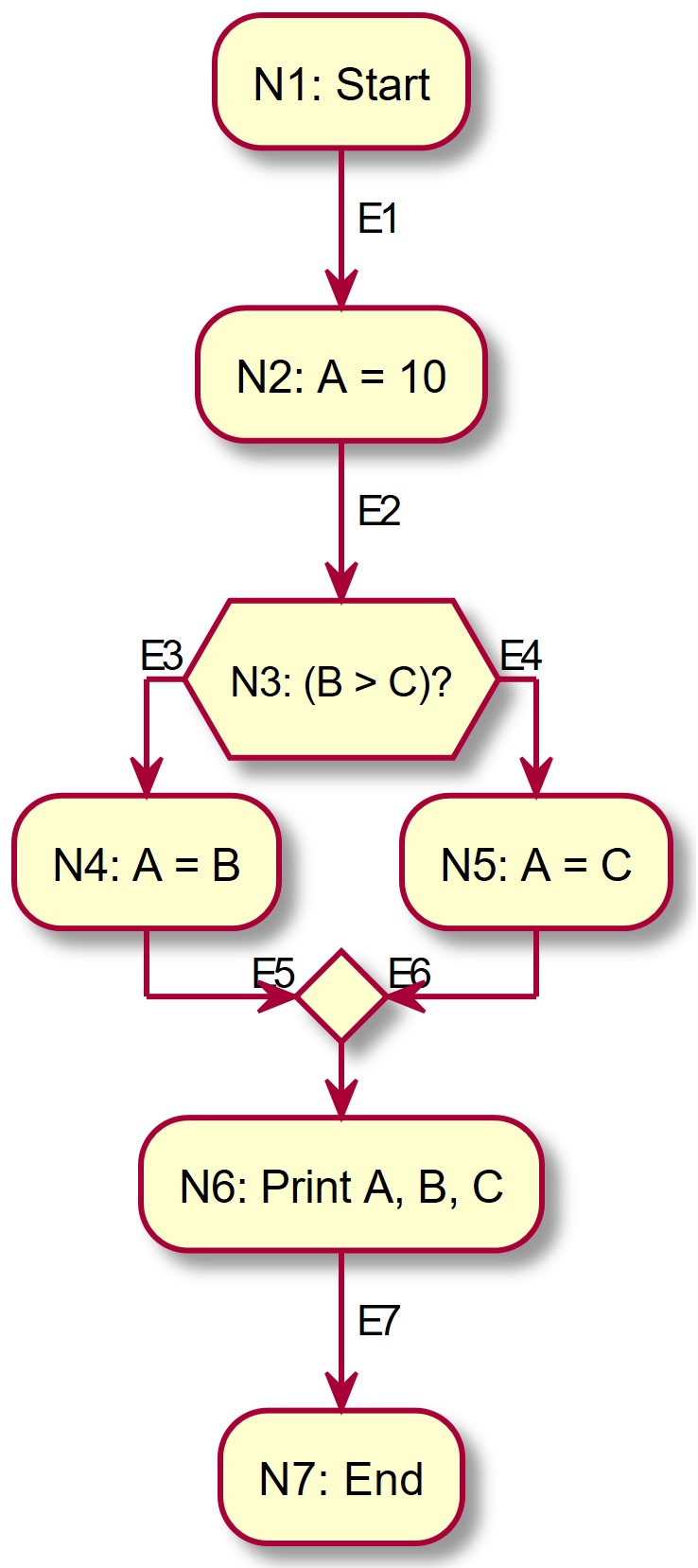}
\caption{UV agent construction metamodel example \cite{sadik2024coding}.} 
\label{fig8}
\end{figure}

In our first experiment, we concentrated on analyzing the structure and complexity of the auto-generated code using the cyclomatic complexity metric. This metric is essential for measuring code complexity, as it enumerates the number of linearly independent paths through the program's source code. The calculation relies on the code's control-flow graph, akin to the example depicted in Fig.~\ref{fig8}. The cyclomatic complexity (\(M\)) is derived from the formula:
\begin{equation}
M = E - N + 2P
\end{equation}
where \(E\) represents the number of edges in the flow graph, \(N\) is the number of nodes, and \(P\) indicates the number of disconnected parts of the graph. For instance, given the graph shown in Fig.~\ref{fig8}, the calculated \(M\) is 3. This \(M\) value is critical for evaluating various software aspects, such as testing difficulty, maintainability, understandability, refactorability, performance, reliability, and documentation quality. Based on the \(M\) value, risk levels are categorized as follows: an \(M\) between 1 and 10 signifies low risk; an \(M\) between 11 and 20 indicates moderate risk; an \(M\) from 21 to 50 suggests high risk, potentially requiring code review or decomposition into smaller modules; and an \(M\) exceeding 50 denotes severe risk, necessitating substantial refactoring.

In our AMDD approach, we placed particular emphasis on the impact of integrating formal constraints on the generation of deployable code. The aim of this experiment was to gauge the influence of the matemodel constraints on the complexity of the auto-generated code. Consequently, we generated two distinct deployments that varied in the degree of constraints embedded within their models. The first deployment was based on a model that implemented only OCL constraints, while the second deployment also integrated FIPA-ontology constraints into the model.

\begin{table}[ht]
\caption{Cyclomatic Complexity Analysis of Agent Classes \cite{sadik2024coding}.}
\label{tab:complexity-analysis}
\centering
\begin{tabular}{|l|l|l|l|l|l|l|l|l|}
\hline
Class & \multicolumn{2}{c|}{Operator} & \multicolumn{2}{c|}{MCC} & \multicolumn{2}{c|}{UVF-Manager} & \multicolumn{2}{c|}{UV} \\ \hline
Constraints & OCL & \begin{tabular}[c]{@{}l@{}}OCL +\\ Ontology\end{tabular} & OCL & \begin{tabular}[c]{@{}l@{}}OCL +\\ Ontology\end{tabular} & OCL & \begin{tabular}[c]{@{}l@{}}OCL +\\ Ontology\end{tabular} & OCL & \begin{tabular}[c]{@{}l@{}}OCL +\\ Ontology\end{tabular} \\ \hline
Edges (E) & 8 & 12 & 15 & 22 & 16 & 23 & 8 & 12 \\ \hline
Nodes (N) & 8 & 11 & 13 & 19 & 14 & 19 & 8 & 11 \\ \hline
Branches (P) & 1 & 1 & 1 & 1 & 1 & 1 & 1 & 1 \\ \hline
Complexity (M) & 2 & 3 & 4 & 5 & 4 & 6 & 2 & 3 \\ \hline
\end{tabular}
\end{table}

Following the generation of these two distinct codebases, we converted the agent classes into control flow diagrams for the purpose of computing their cyclomatic complexity (\(M\)), as summarized in Table 1. A comparison of the \(M\) values from the auto-generated code indicates that complexity incrementally rises with the addition of FIPA-ontology constraints. However, it's noteworthy that the complexity levels for all classes in both deployments remained within the low-risk category. This observation suggests that the structure of the auto-generated code is sufficiently robust and does not require further refactoring. Importantly, the highest observed \(M\) value, which was 6, was associated with the UVF-manager class in the second deployment that included both OCL and FIPA-ontology constraints. This finding indicates a significant margin for incorporating additional constraints into our model without surpassing the low-risk threshold.

\subsection{Behaviour evaluation}
\begin{figure}[h!]
\centering
\includegraphics[width=12cm]{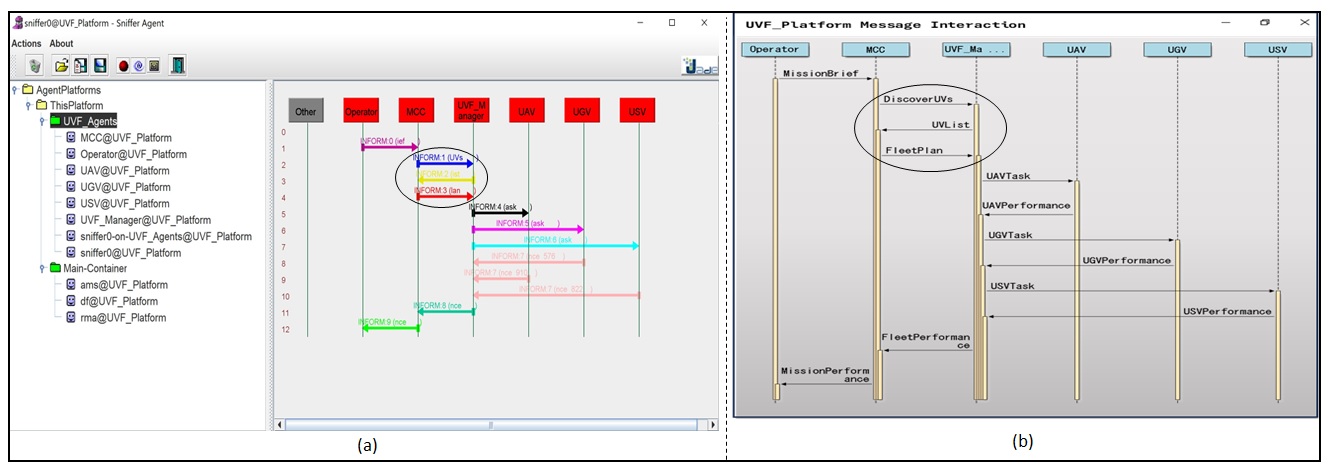}
\caption{UV agent construction metamodel example \cite{sadik2024coding}.} 
\label{fig9}
\end{figure}

In the second experiment of our study, we generated two distinct deployments: one in Java for the JADE platform, and another in Python for the PADE framework. The objective was to assess and compare the behaviors of code executed on JADE with that running on PADE, aiming to verify the system dynamics' consistency across different programming languages. This examination specifically focused on the agents' interaction behaviors within both the JADE and PADE frameworks. Observations indicated that the agents' behaviors, as captured by the JADE Sniffer tool, align with the sequence diagrams generated from PADE agent interactions, as depicted in Fig.~\ref{fig9}.

In the sequence diagrams presented in Fig.~\ref{fig9}, the interaction sequence initiates with the operator sending a mission brief to the MCC. Upon receiving this message, the MCC directs the UVF-manager to identify available UVs. With a list of accessible UVs at hand, the MCC formulates a fleet plan and relays it to the UVF-manager, who subsequently assigns specific tasks to the available UVs. Each UV, upon completing its task, transmits performance data back to the UVF-manager, who then compiles a comprehensive fleet-performance metric. This metric is forwarded to the MCC, which evaluates it in relation to the mission objectives to generate a definitive mission-performance report. This report marks the operation's culmination and is ultimately delivered to the operator.

Two notable observations emerged from the comparison of the sequence diagrams in Fig.~\ref{fig9} with the original case study's activity diagram in Fig.~\ref{fig4}. Firstly, an enhancement in interaction was identified, introducing new behaviors for both the MCC and UVF-manager agents. This enhancement was particularly observed in the MCC dispatching a \textit{DiscoverUVs} message to the UVF-Manager and awaiting the \textit{UVList} before developing a \textit{FleetPlan}, logically necessitating an understanding of the available UV resources prior to their mission-specific allocation. This novel interaction behavior was not explicitly detailed in the original case study's activity diagram. Secondly, variations in the timing of interactions between the MCC and UVs were noted between the JADE and PADE implementations, likely reflecting differences in the state machine of each UV instance. This variance suggests that these state machines accurately emulate the operational dynamics of the agents.

\section{DISCUSSION, CONCLUSION, AND FUTURE WORK}

Our investigation into the integration of LLMs within MDD practices has unveiled significant potential for overcoming the challenges posed by natural language ambiguities in auto-generating deployable code. By employing formal modeling languages, such as UML, and enriching them with precise metamodels'  constraints, we have significantly bridged this gap. Our proposed AMDD approach, which introduces semantic depth through metamodelling, has proven to enhance the accuracy of code generation by LLMs, particularly GPT-4, which demonstrates advanced reasoning capabilities.

In our case study focusing on a MAS of an UVF, we utilized class diagrams to represent agents and employed activity and state diagrams to capture their interactions and behaviors. Detailed metamodels were defined by using the OCL for structural aspects and FIPA-ontology for agent communication. This foundational model facilitated the auto-generation of code in both Java and Python, showcasing the effectiveness of our approach.

Our evaluation through two distinct experiments. The first experiment examined the auto-generated code behavior in Java and Python. Both codes run in Java and Python showed identical behavior with the model activity diagram, however new behavior in both implementations was added by GPT-4 as shown in , to reveal the capability of LLM to interpret and then compliment the   . While the second experiment assessed the structural complexity of the generated code—revealed that the addition of FIPA-ontology metamodel does not unduly increase code complexity. This suggests that our approach allows for the inclusion of further constraints without necessitating immediate code refactoring, indicating robustness and scalability.

The insights gained from integrating LLMs into MDD underscore a transformative path towards achieving the sought-after agility in current MDD practices. Future research will focus on assessing the correctness of auto-generated code, incorporating new privacy and cybersecurity metamodels, and comparing our methodology against existing MDD frameworks. Through continuous enhancement and evaluation, we aim to pave the way for more agile, efficient, and robust software development methodologies. As the domain of software development continues to evolve, the seamless interplay between structured modeling, advanced LLM reasoning, and assessments of structural complexity will be paramount. Our research represents a step forward in this direction, suggesting a future where the generation of deployable code from high-level models becomes more streamlined, precise, and reliable.

%
%
%
%

\bibliographystyle{splncs04}
\bibliography{main}

\end{document}